\documentclass[10pt,twocolumn,letterpaper]{article}

\usepackage[pagenumbers]{cvpr} % To force page numbers, e.g. for an arXiv version
\makeatletter
\@namedef{ver@everyshi.sty}{}
\makeatother

\usepackage{graphicx}
\usepackage{amsmath}
\usepackage{amssymb}
\usepackage{booktabs}
\usepackage{todonotes}
\usepackage{tabularx}
\usepackage{multirow}
\usepackage{multicol}
\usepackage{cite}
\usepackage{bm}
\usepackage{url}
\usepackage{balance}
\usepackage{enumitem}
\usepackage{color}
\usepackage{bbding}
\usepackage{nccmath}
\usepackage{adjustbox}
\usepackage{tikzinclude}

\usepackage[pagebackref,breaklinks,colorlinks]{hyperref}

\usepackage[capitalize]{cleveref}
\crefname{section}{Sec.}{Secs.}
\Crefname{section}{Section}{Sections}
\Crefname{table}{Table}{Tables}
\crefname{table}{Tab.}{Tabs.}

\makeatletter
\def\maketag@@@#1{\hbox{\m@th\normalfont\normalsize#1}}
\makeatother

\def\VEC#1{{\boldsymbol{#1}}}

\def\vL{\VEC{L}}
\def\vN{\VEC{N}}
\def\vV{\VEC{V}}
\def\vH{\VEC{H}}
\def\vn{\VEC{n}}
\def\vp{\VEC{p}}

\begin{document}

%%%%%%%%% TITLE
\title{Fresnel Microfacet BRDF:\\%
Unification of Polari-Radiometric Surface-Body Reflection}

\author{Tomoki Ichikawa \qquad Yoshiki Fukao \qquad Shohei Nobuhara \qquad Ko Nishino\\
Graduate School of Informatics, Kyoto University
}
\date{}

\twocolumn[{
  \maketitle 
  \begin{center}
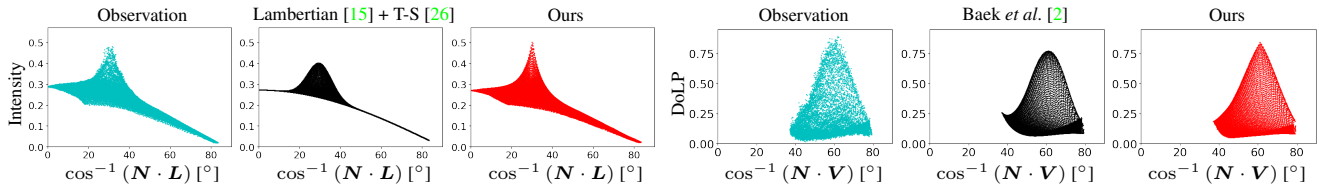

    \captionsetup{type=figure}

    \includetikzgraphics[opening]{fig/tikz}

    \captionof{figure}{We derive Fresnel Microfacet BRDF, a novel physically-based BRDF model which consolidates radiometric (left) and polarimetric (right) reflection as well as body and surface reflections of surface microgeometry in a single model. The model outperforms past physically-based models (\eg, Lambertian \cite{Lambert1760} plus Torrance-Sparrow \cite{torrance1967theory} and pBRDF \cite{Baek2018SimultaneousAO}) in accuracy and generality.}
    \label{fig:opening figure}
\end{center}
}]

%%%%%%%%% ABSTRACT
\begin{abstract}
Computer vision applications have heavily relied on the linear combination of Lambertian diffuse and microfacet specular reflection models for representing reflected radiance, which turns out to be physically incompatible and limited in applicability. In this paper, we derive a novel analytical reflectance model, which we refer to as Fresnel Microfacet BRDF model, that is physically accurate and generalizes to various real-world surfaces. Our key idea is to model the Fresnel reflection and transmission of the surface microgeometry with a collection of oriented mirror facets, both for body and surface reflections. We carefully derive the Fresnel reflection and transmission for each microfacet as well as the light transport between them in the subsurface. This physically-grounded modeling also allows us to express the polarimetric behavior of reflected light in addition to its radiometric behavior. That is, FMBRDF unifies not only body and surface reflections but also light reflection in radiometry and polarization and represents them in a single model. Experimental results demonstrate its effectiveness in accuracy, expressive power, and image-based estimation. 
\end{abstract}

%%%%%%%%% BODY TEXT
\section{Introduction}
\label{sec:introduction}

Reflection is a fundamental physical phenomenon of light that serves as a key creator of our rich visual world. Models of light reflection lie at the heart of visual information processing both for synthesis and analysis. In computer vision, reflectance models play an essential role in 3D reconstruction, inverse rendering, and material estimation. The goal is to invert light reflection to deduce its physical ingredients, such as the surface geometry, from images. Naturally, devising simple yet accurate models that are faithful to the underlying physics becomes vital. Analytical models provide a sound basis for solving these inverse problems as parameter estimation and physically-based models lend semantic interpretations of the results.

Physically-based analytical reflectance models have been studied extensively. Parametric representations of the Bidirectional Reflectance Distribution Function (BRDF) are of particular importance, as they enable pixel-wise estimation of its parameters. Most models widely adopted in computer vision are built on two representative models corresponding to the two distinct reflection components, namely body reflection and surface reflection. Body reflection refers to the light that transmits into the subsurface and is eventually emitted from the surface. It is also called diffuse reflection as it is comparatively scattered in directions. The Lambertian reflectance model \cite{Lambert1760} which models it as uniform distribution in the angular domain dominates computer vision applications due to its simple linear form.

Surface reflection is the light that immediately reflects off the surface. It is also referred to as specular reflection as it primarily concentrates around the perfect mirror reflection direction of incident light. Torrance and Sparrow \cite{torrance1967theory} introduced the idea of modeling the microgeometry within a single pixel that causes this angular spread of surface reflection with oriented mirror microfacets. Then on, many models have built on this key idea of oriented microfacets \cite{cook1982reflectance, Walter2007microfacet}. Oren and Nayar \cite{Oren1995} applied the idea to body reflection by assuming Lambertian instead of mirror microfacets.

A linear combination of these diffuse and specular reflection models, most often Lambertian or Oren-Nayar plus Torrance-Sparrow, have been widely used in vision applications. There are, however, three problems that fundamentally limit the accuracy of such a reflection representation. The first is that the two reflection components are modeled on inconsistent microgeometry. Lambertian and other body reflection models assume a single Lambertian microfacet or an oriented distribution of Lambertian microfacets \cite{Oren1995}, while specular reflection models assume mirror microfacets \cite{torrance1967theory}. This is physically implausible and also hinders physical interpretation of the parameter estimates. 

The second is that past diffuse reflection models do not account for light transport inside the microgeometry. The Oren-Nayar model ignores discrepancies in incident and exitant microfacets. This can be fine for mesoscopic and macroscopic geometry (\ie, Bidirectional Texture Function) as demonstrated in their work \cite{Oren1995}, but leads to significant inaccuracy for microgeometry (\ie, regular imaging conditions). Incident light to one microfacet will likely exit from a different microfacet whose effect cannot be ignored for accurate body reflection representation. 

The third is that estimation of the parameter values (\ie, reflectometry) of such linear combinations of diffuse and specular reflection models is inherently unstable. Specular reflection is usually either sparse (\eg, a shiny surface with a narrow highlight) or weak (\eg, a rough surface with a broad specular lobe). This makes estimation of specular parameter values while disentangling diffuse and specular components challenging. Most works thus require multiple images captured from different imaging conditions.

In this paper, we derive a novel analytical reflectance model that is physically accurate and generalizes to various real-world surfaces. Our key idea is to build up from the very atomic behavior of light reflection, namely Fresnel reflection. We model surface microgeometry with a collection of oriented mirror facets, both for body and surface reflections. We carefully derive the Fresnel reflection and transmission for each microfacet as well as the light transport between them in the subsurface. By modeling the full Fresnel behavior of light for an analytically oriented distribution of mirror microfacets, we arrive at a generalized reflection model that subsumes past representative models as special cases. This physically-grounded modeling allows us to describe the polarimetric behavior of reflected light by a rough surface, in addition to its radiometric behavior. As a result, our novel reflectance model, which we refer to as Fresnel Microfacet BRDF model (FMBRDF), unifies not only body and surface reflections but also light reflection in radiometry and polarization in a single model.

We experimentally validate our FMBRDF model by evaluating its accuracy with a wide range of measured BRDFs and images of real surfaces. The results show that FMBRDF can accurately model both the intensity and polarization, particularly in comparison with past representative models. We also show that FMBRDF can be estimated from a single polarimetric image.

To the best of our knowledge, FMBRDF is the first reflectance model to seamlessly unify body and surface reflections with the same microgeometry and also describe both its radiometric and polarimetric light reflections in a single model. We believe FMBRDF will provide an invaluable basis for accurate radiometric and polarimetric image analysis and serve as a backbone for a wide range of computer vision applications. We will release all the code and data to ease its adoption.

\section{Related Works}

\newcolumntype{C}{>{\centering\arraybackslash}X}
\begin{table}[t]
    \centering
    \scriptsize
    \begin{tabularx}{\linewidth}{cCCCCC}
        \hline
        Model & MSR & MBR & FT & MLT & Pol. \\ \hline
        T-S~\cite{torrance1967theory} + Lambertian & \checkmark & & & & \\
        T-S~\cite{torrance1967theory} + O-N~\cite{Oren1995} & \checkmark & \checkmark & & & \\
        Baek \etal~\cite{Baek2018SimultaneousAO} & \checkmark & & \checkmark & & \checkmark \\
        Ours & \checkmark & \checkmark & \checkmark & \checkmark & \checkmark \\ \hline
    \end{tabularx}
    \caption{Our Fresnel Microfacet BRDF model is, to our knowledge, the first physically-based reflection model that accurately expresses microfacet surface reflection (MSR), microfacet body reflection (MBR), Fresnel transmission (FT), microscopic light transport (MLT), and polarization (Pol.) in a single model.}
    \label{tab:models comparison}
\end{table}

Light reflection at a surface point can be described by the Bidirectional Reflectance Distribution Function (BRDF): the ratio of the reflected surface radiance to the incident irradiance~\cite{Nic77}. Various models have been introduced to approximate the BRDF of real-world surfaces.
\Cref{tab:models comparison} summarizes the fundamental differences between our model and representative physically-based models.

\vspace{-12pt}
\paragraph{Phenomenological Models: }
Various empirical models have been introduced in the past. Phong~\cite{Phong1975IlluminationFC} proposed a specular reflection model based on the power of the angle made by the mirror reflected incident light direction and the viewing direction. Lafortune \etal~\cite{Lafortune1997} generalized the Phong model with multiple specular lobes. Koenderink and van Doorn \cite{Koenderink98phenomenologicaldescription} used Zernike polynomials, Ramamoorthi and Hanrahan \cite{ramamoorthi2001signal} spherical harmonics, and Edwards \etal \cite{edwards2006halfway} 2D Gaussians on halfway disks as the basis to describe a BRDF with basis expansion. These models are phenomenological and do not describe the physical light interaction.

\vspace{-12pt}
\paragraph{Data-driven Models: }
Inductive reflectance models can also be derived by fitting to measurement data. Matusik \etal \cite{matusik2003data} introduced a data-driven non-parametric reflection model using nonlinear dimensionality reduction of 100 measured BRDFs. Nishino \cite{nishino2009directional} introduced the directional statistical BRDF model by viewing reflected light as a hemispherical statistical distribution. Romeiro \etal \cite{romeiro2008passive} applied non-negative matrix factorization to the angular tabulation of BRDFs. More recently, Chen \etal \cite{ChenECCV2020} derived the iBRDF model based on an invertible neural network. These data-driven models are expressive, especially when modeling those BRDFs in the vicinity of the training data, but do not offer physical interpretations of the surface. Their generalizability also solely hinges on the ability to collect dense angular measurements of the actual BRDF.

\vspace{-12pt}
\paragraph{Physically-based Models: }
Deriving an analytical expression of the physical process of light reflection at a surface point has been a long-standing problem. Torrance and Sparrow \cite{torrance1967theory} introduced the idea of modeling the microgeometry of a surface with a distribution of oriented microfacets each of which mirror-reflects light. Cook and Torrance \cite{cook1982reflectance} extended it with the Beckmann distribution to model wavelength-dependency. Walter \etal \cite{Walter2007microfacet} applied the Cook-Torrance model to model refractive transmission of translucent materials.
Oren and Nayar \cite{Oren1995} generalized the Lambertian model by modeling the surface microgeometry with microfacets of purely Lambertian reflection. These models, however, do not model the full BRDF and should not be simply combined (see \cref{sec:introduction}).

\vspace{-12pt}
\paragraph{Polarimetric Reflection Model: }
Polarimetric reflection models have been introduced mainly for geometry recovery from polarization. Atkinson and Hancock \cite{Atkinson2007ShapeEU} assume pure diffuse reflection to estimate the zenith angles of surface normals from observed degrees of polarization.
Others assume either pure diffuse or mirror reflection \cite{ma2007rapid,hyde2009geometrical,Zhu2019polrgbstereo,Smith2018HfP}. Baek \etal \cite{Baek2018SimultaneousAO} introduced a polarimetric BRDF (pBRDF) model consisting of both diffuse and specular reflections, the latter of which is based on microfacet geometry (\ie, a polarimetric version of Lambertian plus Torrance-Sparrow). Extensions of this model with a unpolarized diffuse term \cite{kondo2020accurate} and a single scattering term \cite{hwang2022sparse} have been introduced. These models, however, consider diffuse reflection on a perfectly flat surface, and require special imaging setups (\eg, co-axial imaging \cite{Baek2018SimultaneousAO}) for parameter estimation. Baek \etal \cite{baek2020image} acquired a pBRDF dataset consisting of 25 materials and introduced a data-driven pBRDF model.

\section{Radiometric Fresnel Microfacet BRDF}
\label{sec:microfacet rediometric reflection}

%%%
\begin{figure}[t]
    \centering
    \resizebox{1.0\linewidth}{!}{
        \adjincludegraphics[Clip={{0.2\width} {0.2\height} {0.15\width} {0.35\height}}]{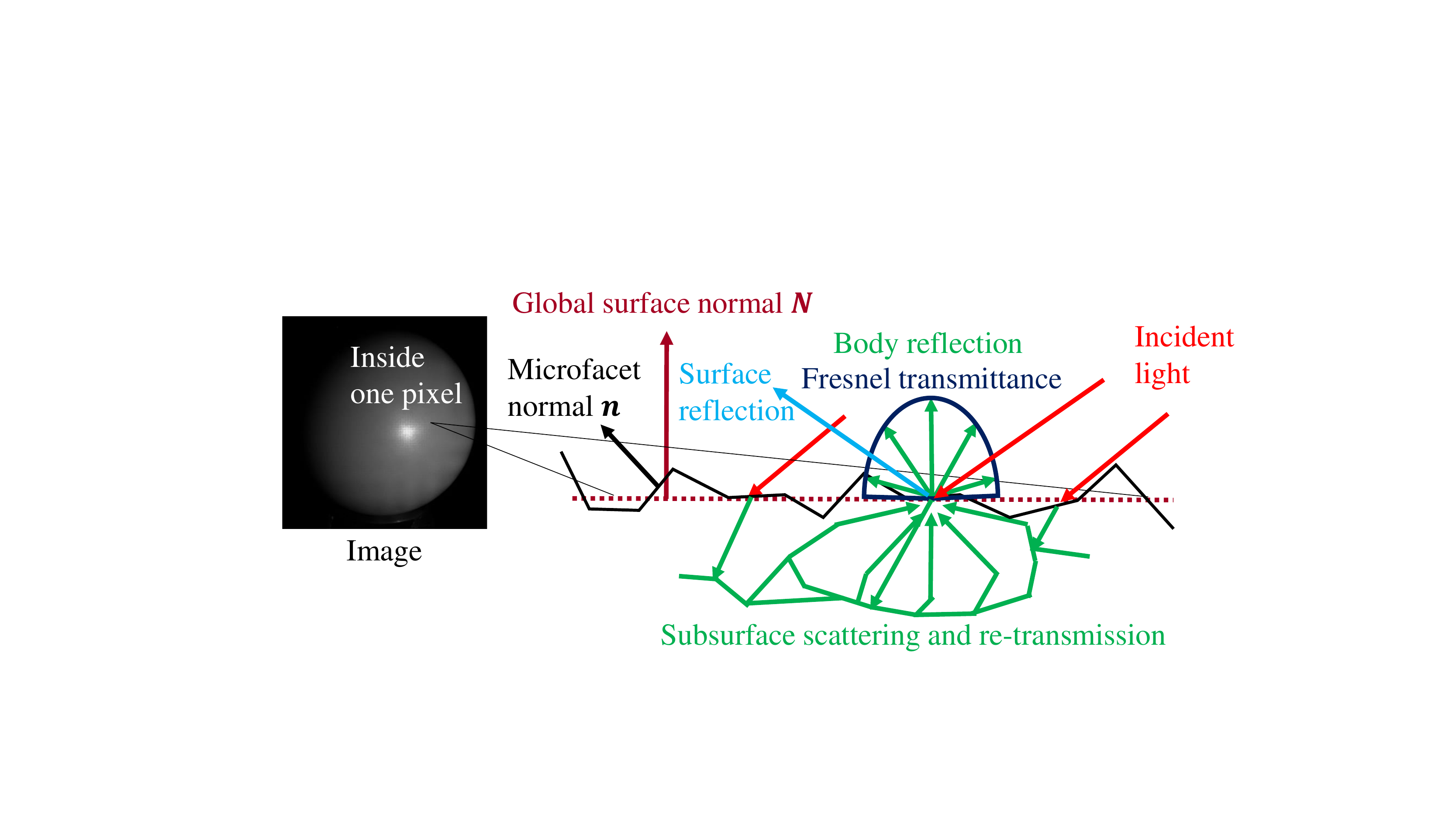}
    }
    \caption{We model the surface microgeometry with a collection of statistically oriented mirror microfacets. Our model accurately captures both the Fresnel reflection (\ie, mirror reflection) and Fresnel transmission and re-transmission (\ie, body reflection) by each microfacet. Light transmitted into the surface body can be re-transmitted from a microfacet different from the incident one after scattering in the body. We also model this light transport inside the microgeometry.}
    
    \label{fig:microfacet}
\end{figure}

We first derive the radiometric FMBRDF, the intensity behavior of our novel reflectance model, by expressing the surface and body reflections caused by the same microgeometry consisting of mirror microfacets.

As depicted in \cref{fig:microfacet}, a single pixel spans an area on the surface. We model the microgeometry in this surface area with a collection of oriented planar microfacets, each of which is perfectly smooth and mirror-reflects light. If we refer to the overall surface normal of this surface area as the global surface normal $\vN$, this means that $\vN$ actually consists of many microfacets each with its own surface normal $\VEC{n}$ oriented according to a statistical distribution. We model the Fresnel reflection and transmittance by each of these mirrored microfacets and express the aggregated reflections by the set of microfacets as surface reflection and body reflection. In contrast, past physically-based models assumed mirrored microfacets for specular reflection and Lambertian microfacets for diffuse reflection, \ie, they had mismatched assumptions on the atomic behavior of individual microfacets.

\subsection{Generalized Radiometric Surface Reflection}

Surface reflection is the aggregated light bundle each of whose constituent light ray is mirror-reflected by the interface of a microfacet. We follow, but significantly extend, the Torrance-Sparrow model \cite{torrance1967theory} to derive an expression that accurately represents this surface reflection based on the fundamental behavior of light reflection, namely Fresnel reflection.

The incident light as a bundle of uni-directional light rays cut by the surface area corresponding to a single pixel is mirror-reflected by microfacets whose normals happen to be aligned with the bisector of the view direction $\vV$ and the incident light direction $\vL$, which is represented by the halfway vector $\vH = (\vL+\vV)/\| \vL+\vV \|$. The aggregated radiance $L_s$ of such reflected light by a distribution of oriented microfacets becomes%~\cite{torrance1967theory}
\begin{equation}
    \label{eq:Torrance-Sparrow}
    L_s = k_s \frac{R(\theta_d)D(\theta_\vH)G(\vH)}{4(\vN\cdot\vV)}E_o \,,
\end{equation}
where $\vN$ is the global surface normal at a pixel, $k_s$ is the surface reflection albedo, $\theta_d$ is incident light angle to the microfacet, $R(\theta_d)$ is the Fresnel reflectance, $\theta_\vH$ is the zenith angle of $\vH$ from $\vN$, $D$ is the microfacet distribution, $G$ is the geometric attenuation factor, and $E_o$ is the irradiance from the light source when the surface normal is aligned with $\vL$ \cite{torrance1967theory}. For simplicity, we omit $\vN$, $\vL$, and $\vV$ from the notations. While $G$ is derived from V-cavity microgeometry in Torrance-Sparrow model \cite{torrance1967theory}, we instead use Smith's masking function and the separable masking-shadowing function \cite{smith1967geometrical, Heitz2014microfacet}.

In the Torrance-Sparrow model, $D$ is a Gaussian and its parameter $\sigma$ represents the surface roughness. We generalize this orientation distribution $D$ by modeling it with the generalized normal distribution, which will be able to express a far wider range of microgeometry
\begin{equation}
  \label{eq:gennormal}
  D(\theta_\vH) \propto \exp\left[ -\left({\theta_\vH}/{\alpha}\right)^\beta \right] \,.
\end{equation} 
Image-based measurement of such a generic microfacet orientation distribution would be impossible with a radiometric BRDF model due to the entanglement of surface and body reflections. We later show that modeling the polarimetric light behavior enables its robust estimation.

\subsection{Generalized Radiometric Body Reflection}
Let us first review foundational models of diffuse reflection, to derive an accurate model of body reflection by mirror microfacets, \ie, the light that first transmits into the microfacet with Fresnel transmittance before reemerging. Lambert expressed the phenomenological body reflection as angularly uniform ``diffuse'' reflection. Oren and Nayar \cite{Oren1995} extended this by devising a Gaussian microfacet orientation distribution similar to the Torrance-Sparrow model, but by assuming each microfacet to realize pure Lambertian reflection. In this model, the radiance observed from a surface patch is the average of the projected radiance of the microfacets weighted by the slope-area distribution. The projected radiance $L_{rp}$ of each microfacet is defined as
\begin{equation}
    L_{rp}(\vn) = \frac{d\Phi_r(\vn)}{(\vn\cdot\vN)dn (\vN\cdot\vV) d\omega_r}\,,
    \label{eq:L_rp O-N}
\end{equation}
where $\vn$ is the normal of the microfacet, $d\Phi_r(\vn)$ is the flux in the viewer direction, $dn$ is the microfacet area, and $d\omega_r$ is an infinitesimal solid angle to the viewer direction. The reflected flux $d\Phi_r(\vn)$ is given by Lambertian reflection.

The projected radiance is weighted by the slope-area distribution, \ie, the foreshortened microfacet areas
\begin{equation}
    L_{b} = \int_{\Omega} L_{rp}(\vn) D(\theta_\vn)\cos\theta_\vn d\omega_\vn \,,
    \label{eq:L_b}
\end{equation}
where $\theta_\vn$ is the zenith angle of $\vn$ from the global surface normal $\vN$, $\Omega$ is the upper hemisphere around $\vN$, $d\omega_\vn$ is an infinitesimal solid angle subtended by the microfacet normal direction. The slope-area distribution is normalized by
\begin{equation}
    \int_{\Omega} D(\theta_\vn)\cos\theta_\vn d\omega_\vn = 1 \,.
\end{equation}
In the Oren-Nayar model, the slope-area distribution is a Gaussian function. 

The Oren-Nayar model, however, has a few shortcomings as a body reflection model as we reviewed in \cref{sec:introduction}. Its key assumption of Lambertian microfacets is fundamentally incompatible with specular reflection models which rightly assume mirror microfacets. In fact, it is not clear how a Lambertian microfacet can be realized in the real world, as Lambertian reflection can only result from rough surfaces (\ie, it cannot be an atomic behavior of light). The original work \cite{Oren1995} fits the Oren-Nayar model to mean intensity values of a large surface patch similar to celestial imaging conditions (\eg, lunar photography), effectively demonstrating its use as a Bidirectional Texture Function (BTF) \cite{DanaBTF} not a BRDF. In other words, it models the reflectance of meso/macrogeometry by assuming Lambertian microgeometry, not individual microfacets. This renders its use for BRDF modeling and per-pixel inverse rendering inappropriate for ordinary imaging conditions. Furthermore, it does not model the light transport between microfacets. This is actually essential to realize Lambertian reflection.

Let us now derive our body reflection model based on the exact same microgeometry underlying the surface reflection. Unlike the Oren-Nayar derivation, we must consider all combinations of incident and outgoing points within the microgeometry of a pixel $A_p$ to accommodate light transport between microfacets. We first focus on a point $\vp$ on an oriented microfacet of normal $\vn$ in $A_p$. We can rewrite $L_{rp}(\vn)$ with the projected radiance $L_{rp}(\vp;\vn)$ at $\vp$ as
\begin{equation}
    L_{rp}(\vn) = \frac{1}{|A_\vn|} \int_{A_\vn} G_m^{b}(\vp;\vn) L_{rp}(\vp;\vn)dA_\vn \,,
    \label{eq:L_rp n}
\end{equation}
where $G_m^{b}(\vp;\vn)$ is a binary masking function at $\vp$, $A_\vn$ is the region of microfacets oriented to $\vn$ in $A_p$, $|\cdot|$ denotes the area of a region, and $dA_\vn$ is an infinitesimal area in the vicinity of $\vp$. The microfacets area $|A_\vn|$ becomes 
\begin{equation}
    |A_\vn| = |A_p|D(\theta_\vn)d\omega_\vn \,.
    \label{eq:A_n}
\end{equation}

Similar to \cref{eq:L_rp O-N}, $L_{rp}(\vp;\vn)$ is defined by
\begin{equation}
    L_{rp}(\vp;\vn) = \frac{d\Phi_r(\vp;\vn)}{(\vn\cdot\vN) dA_\vn (\vN\cdot\vV) d\omega_r} \,,
    \label{eq:L_rp pn}
\end{equation}
where $d\Phi_r(\vp;\vn)$ is the flux in the viewer direction at point $\vp$ and $\omega_r$ is an infinitesimal solid angle to the viewer direction. The flux $d\Phi_r(\vp;\vn)$ is represented with the radiance $L_{r}(\vp;\vn)$ to the viewer direction at $\vp$ as
\begin{equation}
    d\Phi_r(\vp;\vn) = L_{r}(\vp;\vn) \cdot (\vn\cdot\vV)dA_\vn d\omega_r \,.
    \label{eq:dPhi_r pn}
\end{equation}

The outgoing light $L_r(\vp;\vn)$ at a point $\vp$ is the sum of the light transported from other incident points on the microgeometry. We define $dL_{r}(\vp, \vp_i;\vn, \vn_i)$ as the radiance of light transported from an incident point $\vp_i$ on a microfacet oriented to $\vn_i$. Assuming that the surface is opaque enough so that no light is transported from outside of the patch $A_p$, we can express $dL_{r}(\vp;\vn)$ with $dL_{r}(\vp, \vp_i;\vn, \vn_i)$ as
\begin{equation}
    dL_{r}(\vp;\vn) = \int_{\Omega} \int_{A_{\vn_i}} \frac{dL_{r}(\vp,\vp_i;\vn,\vn_i)}{d\omega_{\vn_i} dA_{\vn_i}} dA_{\vn_i} d\omega_{\vn_i} \,,
    \label{eq:dL_r pn}
\end{equation}
where $d\omega_{\vn_i}$ is an infinitesimal solid angle that represents the normal direction $\vn_i$ of an incident microfacet, $dA_{\vn_i}$ is an infinitesimal area in the vicinity of $\vp_i$, $A_{\vn_i}$ is the region of microfacets oriented to $\vn_i$ in $A_p$, and its area $|A_{\vn_i}|$ is obtained by
\begin{equation}
    |A_{\vn_i}| = |A_p|D(\theta_{\vn_i})d\omega_{\vn_i} \,.
    \label{eq:A_ni}
\end{equation}

The incident flux $\Phi_i(\vp_i;\vn_i)$ at $\vp_i$ is scattered and distributed to other points on other microfacets near $\vp_i$. To represent outgoing radiance $dL_{r}(\vp, \vp_i;\vn, \vn_i)$ with the incident flux $d\Phi_i(\vp_i;\vn_i)$ at $\vp_i$, we define $S(\vp,\vp_i;\vn, \vn_i)$ as
\begin{equation}
    S(\vp,\vp_i;\vn,\vn_i) = \frac{dL_r(\vp,\vp_i;\vn,\vn_i)}{d\Phi_i(\vp_i;\vn_i)} \,.
    \label{eq:transmission and scattering function}
\end{equation}
This function is similar to but different from the Bidirectional Subsurface Scattering Reflectance Distribution Function (BSSRDF) \cite{Nic77} as the normals at $\vp$ and $\vp_i$ are different. It describes how much light is transmitted into the surface, transported from $\vp_i$ to $\vp$, and re-transmitted into the air.
We can decompose it into transmission, light transport, and re-transmission terms and obtain
\begin{equation}
    S(\vp,\vp_i;\vn,\vn_i) = T(\theta_o) S'(\vp,\vp_i;\vn,\vn_i) T(\theta_i) \,,
    \label{eq:decomposition of S}
\end{equation}
where $T(\theta)$ is the Fresnel transmittance for unpolarized light from the air at incident angle $\theta$, $\theta_i = \cos^{-1}(\vn_i \cdot\vL)$, $\theta_o = \cos^{-1}(\vn\cdot\vV)$, and $S'(\vp,\vp_i;\vn,\vn_i)$ represents scattering and absorption.

When the light source direction is $\vL$, we obtain the incident flux $d\Phi_i(\vp_i;\vn_i)$ as
\begin{equation}
    d\Phi_i(\vp_i;\vn_i) = G_s^{b}(\vp_i;\vn_i) \cdot (\vn_i\cdot\vL) E_o dA_{\vn_i} \,,
    \label{eq:Phi_i}
\end{equation}
where $G_s^{b}(\vp_i;\vn_i)$ is a binary shadowing function at $\vp_i$.
From \cref{eq:L_rp n,eq:L_rp pn,eq:dPhi_r pn,eq:dL_r pn,eq:A_ni,eq:transmission and scattering function,eq:decomposition of S,eq:Phi_i} the projected radiance $L_{rp}(\vn)$ becomes
\begingroup\makeatletter\def\f@size{8}\check@mathfonts
\begin{align}
    \begin{split}
        L_{rp}(\vn) &= \frac{\vn\cdot\vV}{(\vn\cdot\vN)(\vN\cdot\vV)} T(\theta_o) \\
        & \int_{\Omega} M(\vn,\vn_i) T(\theta_i) (\vL\cdot\vn_i) |A_p|D(\theta_{\vn_i})d\omega_{n_i} \cdot E_o \,,
    \end{split}
    \label{eq:L_rp n2}
\end{align}
\endgroup
where
\begingroup\makeatletter\def\f@size{8}\check@mathfonts
\begin{align}
    \begin{split}
    & M(\vn,\vn_i) = \\ & \frac{1}{|A_\vn| |A_{\vn_i}|} \int_{A_\vn}\!\int_{A_{\vn_i}}\!\!\!\!\!\! G_m^{b}(\vp) G_s^{b}(\vp_i) S'(\vp,\vp_i)dA_{\vn_i}dA_\vn \,.
    \end{split}
    \label{eq:M nni}
\end{align}
\endgroup
For simplicity, we omit $\vn$ and $\vn_i$ from the notations for $G_m^{b}$, $G_s^{b}$, and $S'$.
From \cref{eq:M nni}, $M(\vn,\vn_i)$ denotes the average of the product of $G_m^{b}(\vp)$, $G_s^{b}(\vp_i)$ for all combinations of light transport $S'(\vp,\vp_i)$ between surface points $\vp$ and $\vp_i$ on microfacets oriented to $\vn$ and $\vn_i$ respectively.

%%%
Let us analyze in detail the masking and shadowing term $M(\vn,\vn_i)$ in \cref{eq:M nni} that takes into account light transport within the microgeometry $S'(\vp,\vp_i)$. The product of the masking and shadowing function $G^b_m(\vp)G^b_s(\vp_i)$ has been studied as joint masking-shadowing functions \cite{Heitz2014microfacet} when $\vp = \vp_i$. We follow the widely used separable masking-shadowing function which we use for the surface reflection model and extend it to the case of $\vp \neq \vp_i$. The separable masking-shadowing model assumes that $G^b_m(\vp)$ and $G^b_s(\vp)$ are independent for $\vp$. Extending this assumption, we assume $G^b_m(\vp)$ and $G^b_s(\vp_i)$ are independent of the distance between $\vp$ and $\vp_i$. From this assumption, the masking-shadowing function $G_m^{b}(\vp)G_s^{b}(\vp_i)$ is independent of the distance between $\vp$ and $\vp_i$. Since the light transport $S'(\vp,\vp_i)$ largely depends on the distance between $\vp$ and $\vp_i$, we can assume that $G^b_m(\vp)G^b_s(\vp_i)$ and $S'(\vp,\vp_i)$ have no correlation for $\vp$ and $\vp_i$. The average of their product in \cref{eq:M nni} becomes the product of each average due to statistical independence between $G^b_m(\vp)G^b_s(\vp_i)$ and $S'(\vp,\vp_i)$, and that of $G^b_m(\vp)$ and $G^b_s(\vp_i)$. We obtain
\begin{equation}
    M(\vn,\vn_i) = G_m(\vn)G_s(\vn_i)\overline{S'}(\vn,\vn_i) \,,
    \label{eq:M nni2}
\end{equation}
where $G_m(\vn)$, $G_s(\vn_i)$, and $\overline{S'}(\vn,\vn_i)$ are the average of $G^b_m(\vp)$, $G^b_s(\vp_i)$, and $S'(\vp,\vp_i)$ for $\vp$ and $\vp_i$, respectively.
For consistency with surface reflection, we use Smith's masking-shadowing function for $G_m(\vn)$ and $G_s(\vn_i)$~\cite{smith1967geometrical, Heitz2014microfacet}.

We analyze $\overline{S'}(\vn,\vn_i)$ which is the average of light transport incident on $\vp_i$ from $\vL$ and outgoing from $\vp$ to $\vV$ under the surface for all combinations of $\vp$ and $\vp_i$. Light transported from $\vp_i$ to $\vp$ may be directional depending on the direction from $\vp_i$ to $\vp$. However, when the direction from $\vp_i$ to $\vp$ is random for $\vp$ in $A_\vn$ and $\vp_i$ in $A_{\vn_i}$, the average light is not directional. Moreover, sufficient scattering weakens the effect of the incident direction $\vL$. We can thus safely assume that $\overline{S'}(\vn,\vn_i)$ does not depend on $\vL$ and $\vV$.

The light transport $\overline{S'}(\vn,\vn_i)$ follows the energy conservation law.
Since $\overline{S'}(\vn,\vn_i)$ does not depend on $\vL$ and $\vV$, energy conservation law becomes
\begin{equation}
    \pi\int_\Omega \overline{S'}(\vn,\vn_i) \cdot \frac{|A_\vn|}{d\omega_\vn}d\omega_\vn = k_b \,,
    \label{eq:energy conservation}
\end{equation}
where $k_b$ is the body reflection albedo, $\pi$ is the sum of energy to all directions, and the integration represents the sum of energy at all points. We assume that $k_b$ is constant for the surface patch.

Let us introduce a microfacet correlation function $f(\vn,\vn_i)$ to interpret \cref{eq:energy conservation} 
\begin{equation}
    \overline{S'}(\vn,\vn_i) = \frac{k_b}{|A_p|\pi}f(\vn,\vn_i) \,.
    \label{eq:microfacet correlation function}
\end{equation}
From \cref{eq:A_n,eq:energy conservation,eq:microfacet correlation function}, $f(\vn,\vn_i)$ is normalized by
\begin{equation}
    \int_\Omega f(\vn,\vn_i)D(\theta_\vn)d\omega_\vn = 1 \,.
    \label{eq:energy conservation2}
\end{equation}
The microfacet correlation function $f(\vn,\vn_i)$ encodes the spatial bias of the microfacet orientation distribution. For example, if most microfacets oriented near $\vn_i$ are located near microfacets with $\vn_i$, $f(\vn,\vn_i)$ becomes an angularly narrow distribution function. In the case of randomly distributed microfacets, $f(\vn,\vn_i)$ become the uniform distribution function.
Since the biased microfacet distributions compose the mesoscopic surface, $f(\vn,\vn_i)$ also represents the mesoscopic surface geometry. When $f(\vn,\vn_i)$ is the uniform distribution, the incident light is distributed to microfacets oriented in any direction uniformly and reflected light is hardly directional, \ie the gross body reflection is similar to Lambertian. On the other hand, when $f(\vn,\vn_i)$ is an angularly narrow distribution, the incident light is distributed within the same mesogeometry and the reflected light is directional to the incident direction, \ie it behaves similarly to the Oren-Nayar model.

We model the microfacet correlation function with a 3D von Mises–Fisher distribution 
\begin{equation}
    f(\vn,\vn_i;\kappa) \propto \exp \left[\kappa (\vn_i\cdot\vn) \right] \,,
\end{equation}
where $\kappa$ is the concentration parameter.

From \cref{eq:L_b,eq:L_rp n2,eq:M nni2,eq:microfacet correlation function}, we obtain the body reflection for the same general microgeometry of surface reflection (\cref{eq:Torrance-Sparrow,eq:gennormal})
\begingroup\makeatletter\def\f@size{8}\check@mathfonts
\begin{align}
    & L_b(\vN,\vL,\vV) = %\nonumber\\
    \frac{1}{\vN\cdot\vV} \int_\Omega \int_\Omega G_m(\vn) D(\theta_\vn) T(\theta_o) (\vV\cdot\vn) \cdot \nonumber\\
    & \frac{k_b}{\pi}f(\vn,\vn_i) \cdot G_s(\vn_i) D(\theta_{\vn_i})T(\theta_i)(\vL\cdot\vn_i) d\omega_\vn d\omega_{\vn_i} \cdot E_o \,.
    \label{eq:L_b final}
\end{align}
\endgroup

\section{Polarimetric Fresnel Microfacet BRDF}
Let us now extend the radiometric model to a full polarimetric model. By carefully deriving the Fresnel reflection and transmittance on microgeometry, we show that the polarimetric behavior embodies distinct signatures of the underlying microgeometry in its reflection.

\subsection{Polarization}
Within the temporal span of an observation (\eg, image exposure), the observed light may consist of a collection of linearly polarized light of varying magnitudes, \ie, elliptically distributed polarization. If we capture this partially polarized light with a polarization filter, the observed intensity becomes a function of the filter angle $\phi_c$:
\begin{equation}
    I(\phi_c) = \overline{I} + \rho\overline{I}\cos{(2\phi_c - 2\phi)} \,,
\end{equation}
where $I_{\mathrm{max}}$ and $I_{\mathrm{min}}$ are the intensities in the major and minor axes of the ellipse, $\overline{I}$ is the average intensity $(=\frac{I_{\mathrm{max}}+I_{\mathrm{min}}}{2})$ and $\rho =\frac{I_{\mathrm{max}}-I_{\mathrm{min}}}{I_{\mathrm{max}}+I_{\mathrm{min}}}$ is the degree of linear polarization (DoLP). The angle $\phi$ is called the angle of linear polarization (AoLP) which is the angle the major axis of the ellipse makes in the image plane.

When light is reflected at a surface point, the light either mirror reflects or transmits into the surface. This can be described by the Fresnel equations, \ie, Fresnel reflectance and transmittance. When light is mirror reflected, it is linearly polarized in the direction perpendicular to the plane of reflection (\ie, s-polarized).
Similarly, when light is transmitted into the surface, it is linearly polarized in the direction parallel to the plane of reflection (\ie, p-polarized).

Let us express these polarization behaviors with Stokes vectors which succinctly summarize polarization states.
In the case of linear polarization, the Stokes vector is computed from the intensity at filter angles $0, \pi/4, \pi/2$, and $3/4\pi$,
\begin{equation}
    \VEC{s} = \footnotesize \begin{bmatrix} s_0 \\ s_1 \\ s_2 \\ 0 \end{bmatrix} = {\footnotesize \begin{bmatrix}
    I(0) + I(\frac{\pi}{2}) \\ I(0) - I(\frac{\pi}{2}) \\ I(\frac{\pi}{4}) - I(\frac{3}{4}\pi) \\ 0
    \end{bmatrix}} = {\footnotesize \begin{bmatrix} 2\overline{I} \\ 2\overline{I}\rho\cos\phi \\ 2\overline{I}\rho\sin\phi \\ 0 \end{bmatrix}}\,.
\end{equation}
The DoLP can be extracted from the Stokes vector
\begin{equation}
    \rho = \frac{\sqrt{s_1^2 + s_2^2}}{s_0} \,.
\end{equation}
We do not model circular polarization as it cannot be measured easily.

The polarization transform by reflection and transmittance is expressed with their corresponding Mueller matrix $\VEC{M}$
\begin{equation}
    \VEC{s}_o = \VEC{M}\VEC{s}_i \,,
\end{equation}
where $\VEC{s}_i$ is the Stokes vector of incident light and $\VEC{s}_o$ is the Stokes vector of reflected or transmitted light.

To express the polarization direction, we need to define the coordinate system of incident light (incident coordinate system) and reflected light (outgoing coordinate system). Since the z-axis is aligned with the direction of the light propagation, we define the z-axes of the incident and outgoing coordinate systems as $\vL$ and $-\vV$, respectively. The x-axis and y-axis of each coordinate system can be defined arbitrarily for different purposes. We denote them as $\VEC{x}_{\{i,o\}}$ and $\VEC{y}_{\{i,o\}}$, respectively.

\begin{figure}[t]
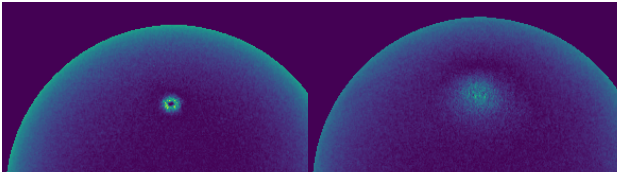

    \centering
    \includetikzgraphics[polintuition]{fig/tikz}

    \caption{Our FMBRDF provides intuitive interpretations of surface reflection polarization, particularly of its degree of polarization. See text for details.}
    
    \label{fig:polintuition}
\end{figure}

\subsection{Polarimetric Microfacet Surface Reflection}
%%%
Let us express the polarimetric microfacet surface reflection of FMBRDF with a Stokes vector and Mueller matrix by rewriting \cref{eq:Torrance-Sparrow}. The Mueller matrix of surface reflection $\VEC{R}$ is given by
\begin{equation}
    \VEC{R}(\theta) = {\footnotesize \begin{bmatrix}
    R_+ & R_- & 0 & 0 \\
    R_- & R_+ & 0 & 0 \\
    0 & 0 & R_\times\cos\delta & 0 \\
    0 & 0 & 0 & R_\times\cos\delta
    \end{bmatrix}} \,,
\end{equation}
where $R_{\pm} = \frac{R_s \pm R_p}{2}$, $R_\times = \sqrt{R_s R_p}$, $\theta$ is the incident light angle, and $\cos\delta$ is $-1$ when $\theta$ is less than the Brewster's angle and $1$ otherwise. $R_s$ and $R_p$ are the Fresnel coefficients 
\begin{equation}
\footnotesize
    R_s(\theta) = \left(\frac{\cos\theta - \mu\cos\theta_t}{\cos\theta + \mu\cos\theta_t}\right)^2\,, \quad
    R_p(\theta) = \left( \frac{\mu\cos\theta - \cos\theta_t}{\mu\cos\theta + \cos\theta_t}\right)^2,
\end{equation}
where $\mu$ is the index of refraction of the object material, and $\theta_t$ is given by Snell's law $\theta_t = \sin^{-1}\left( \frac{1}{\mu}\sin\theta \right)$.

For surface reflection, we can follow the polarimetric specular BRDF model by Baek \etal \cite{Baek2018SimultaneousAO}. To consider polarization, the Fresnel reflectance in \cref{eq:Torrance-Sparrow} is replaced with the Mueller matrix of surface reflection $\VEC{R}$ and a rotation matrix $\VEC{C}$ that represents the polarization direction:
\begin{equation}
    \VEC{C}(\varphi) = {\footnotesize \begin{bmatrix}
    1 & 0 & 0 & 0 \\
    0 & \cos(2\varphi) & -\sin(2\varphi) & 0 \\
    0 & \sin(2\varphi) & \cos(2\varphi) & 0 \\
    0 & 0 & 0 & 1
    \end{bmatrix}} \,.
\end{equation}
The Stokes vector of surface reflection $\VEC{s}_s$ becomes
\begin{equation}
    \VEC{s}_s = k_s\frac{D(\theta_\vH)G(\vH)}{4(\vN\cdot\vV)} \VEC{C}(\varphi_{o,s})\VEC{R}(\theta_d)\VEC{C}(-\varphi_{i,s}) \VEC{s}_i \,,
\end{equation}
where $\varphi_{\{i,o\},s}$ is the angle between the $\VEC{y}_{\{i,o\}}$ and the microfacet normal $\VEC{H}$ projected onto $\VEC{x}_{\{i,o\}}\VEC{y}_{\{i,o\}}$-plane, and $\VEC{s}_i$ is the Stokes vector of the incident light.

\subsection{Polarimetric Microfacet Body Reflection}
%%%
We need to express the body reflection \cref{eq:L_b final} as the aggregated retransmitted light from the oriented mirror microfacets with a Stokes vector and Mueller matrix. The polarization is transformed by the transmission, scattering in the subsurface, and then the retransmission into air.
Since subsurface scattering depolarizes transmitted light, the Mueller matrix is thus given by
\begin{equation}
    \VEC{D}_p\left(\frac{k_b}{\pi}\right) = {\footnotesize \begin{bmatrix}
    \frac{k_b}{\pi} & 0 & 0 & 0 \\
    0 & 0 & 0 & 0 \\
    0 & 0 & 0 & 0 \\
    0 & 0 & 0 & 0 
    \end{bmatrix}} \,.
\end{equation}
The polarization transform by Fresnel transmittance is expressed with the Fresnel coefficients
\begin{equation}
    \VEC{T}(\theta) = {\footnotesize \begin{bmatrix}
    T_+ & T_- & 0 & 0 \\
    T_- & T_+ & 0 & 0 \\
    0 & 0 & T_\times & 0 \\
    0 & 0 & 0 & T_\times
    \end{bmatrix}} \,,
\end{equation}
where $T_{\pm} = \frac{T_s \pm T_p}{2}$, $T_\times = \sqrt{T_s T_p}$, and $\theta$ is the incident light angle. $T_s$ and $T_p$ are the Fresnel coefficients 
\begin{equation}
    T_s(\theta) = 1 - R_s(\theta) \,, \quad \\
    T_p(\theta) = 1 - R_p(\theta) \,.
\end{equation}
Replacing $T(\theta_o)$, $T(\theta_i)$, and $\frac{k_b}{\pi}$ in \cref{eq:L_b final} with $\VEC{C}(\varphi_{o,b})\VEC{T}(\theta_o)$, $\VEC{T}(\theta_i)\VEC{C}(\varphi_{i,b})$, and $\VEC{D}_p\left( \frac{k_b}{\pi} \right)$, the Stokes vector of body reflection $\VEC{s}_b$ becomes
\begingroup\makeatletter\def\f@size{8}\check@mathfonts
\begin{align}
    & \VEC{s}_b = \frac{1}{\vN\cdot\vV} \int_\Omega \int_\Omega G_m(\vn) D(\theta_\vn) (\vV\cdot\vn) \VEC{C}(\varphi_{o,b})\VEC{T}(\theta_o) \VEC{D}_p\left(\frac{k_b}{\pi}\right) \nonumber\\
    & f(\vn,\vn_i) \cdot G_s(\vn_i) D(\theta_{\vn_i})(\vL\cdot\vn_i) \VEC{T}(\theta_i)\VEC{C}(\varphi_{i,b}) d\omega_\vn d\omega_{\vn_i} \cdot \VEC{s}_i \,,
    \label{eq:observed body reflection Stokes vector}
\end{align}
\endgroup
where $\varphi_{\{i,o\},b}$ is the angle between $\VEC{y}_{\{i,o\}}$ and $\vn_i$ and $\vn$ projected onto the $\VEC{x}_{\{i,o\}}\VEC{y}_{\{i,o\}}$-plane.

\subsection{Polarimetric Interpretation}

As our FMBRDF model is physically-based, it provides an intuitive interpretation of polarization of surface reflection. Polarimetric image interpretation is notoriously difficult but our model lets us map key characteristics to its physically explicable parameters. As shown in \cref{fig:polintuition}, let us consider the degree of linear polarization (DoLP) of a sphere, \ie, an object that has all possible frontal facing surface normals. The roughness and shape of the microfacet orientation distribution function dictate the DoLP distribution on the surface. The further the surface normal is from the half vector, the more the observed surface reflection light attenuates and the smaller the DoLP as body reflection dampens the polarization by surface reflection. This relation of the DoLP and the angle between the half vector and surface normal $\theta_{\vH, \vn}$ embodies the shape $\beta$ of the distribution function. The roughness is encoded in the width of this DoLP lobe for $\theta_{\vH, \vn}$. The index of refraction determines the Fresnel transmittance, which in turn governs the DoLP of the region dominated by body reflection. The albedo ratio $r_k=\frac{k_b}{k_s}$ determines the scale of the DoLP lobe for $\theta_{\vH, \vn}$. The microfacet correlation parameter $\kappa$ controls the DoLP of body reflection in surface areas whose global surface normal is aligned with the lighting direction. For instance, when $\kappa$ is large, the polarization by the microfacets aligned with the global surface normal is dampened less by light from other microfacets.

\begin{figure}[t]
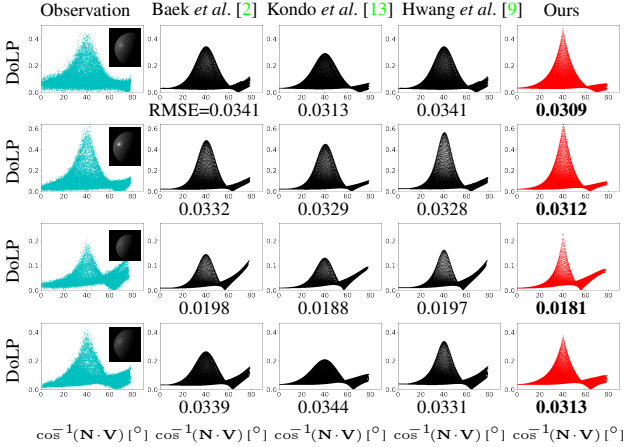

\centering

    \includetikzgraphics[polaccuracy]{fig/tikz}
    
    \caption{Polarimetric accuracy shown with DoLP values plotted as a function of the angle between the global surface normal and the viewing direction. Each graph shows the fitting results for one of the lighting conditions. The number under the graph is the root mean square error. Our FMBRDF model accurately captures the characteristics of the DoLP distributions both for surface and body reflections regardless of the surface roughness and color.}
    \label{fig:BRDFestimation}
\end{figure}

\section{Polari-Radiometric Reflectometry}

The newly derived FMBRDF describes both the polarimetric and radiometric behaviors of light reflected by a surface. This allows us to estimate its parameters from a single polarimetric image captured with a known directional light of an object of known geometry and then use those parameters to analyze and synthesize not just polarimetric but also radiometric appearance of that surface. 

Given a single polarimetric image, the parameters of FMBRDF, namely the index of refraction $\mu$, albedo ratio $r_k=\frac{k_b}{k_s}$, surface reflection albedo $k_s$, surface roughness $\alpha$, and shape of orientation distribution $\beta$, can be estimated with least squares optimization 
\begin{equation}
    \min_{\mu, r_k, k_s, \alpha, \beta, \kappa} \frac{1}{M}\sum_i^M(\overline{I}_i - I_i)^2 + \frac{\sum_i^M w_i(\overline{\rho}_i-\rho_i)^2}{\sum_i^M w_i}\,,
    \label{eq:optimization}
\end{equation}
where $\overline{\rho}_i$ and $\rho_i$ are the observed and rendered (\ie, computed with FMBRDF) DoLPs at pixel $i$, respectively, $\overline{I}_i$ and $I_i$ are the observed and rendered intensity at pixel $i$, respectively, and $M$ is the number of pixels. Note that only the intensity and DoLP values are necessary, and the use of the angle of linear polarization (AoLP) is avoided as their measurements are usually noisy. The weight $w_i$ is set to the ratio of the number of inliers and outliers when pixel $i$ is an outlier, and 1 otherwise, to account for the imbalance in the surface areas dominated with surface or body reflection. We avoid the costly integration of body reflection and Smith's masking function by approximating each with a simple multi-layer perceptron (MLP) and leverage their fast inference and differentiability. We solve the non-linear optimization using Adam \cite{DBLP:journals/corr/KingmaB14}.

\section{Experimental Results}
\begin{figure}[t]
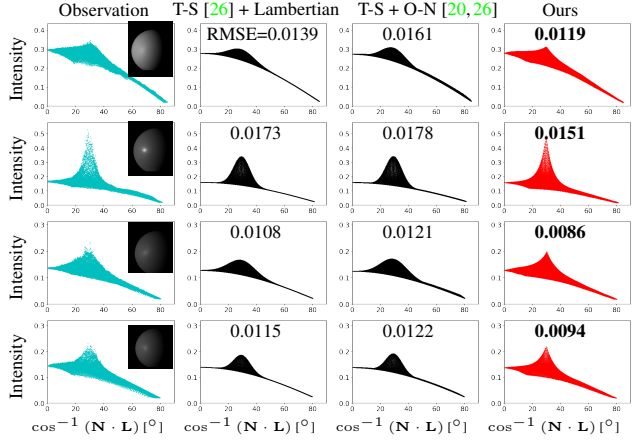

\centering
    \includetikzgraphics[radaccuracy]{fig/tikz}    

    \caption{Radiometric accuracy shown with intensity values as a function of the angle between the global surface normal and the light source direction. Each graph shows the fitting results for one of the lighting conditions. The number at the top of the graph is the RMSE. Our FMBRDF model accurately represents both surface and body reflections regardless of the surface roughness and color.}
    \label{fig:radiometric accuracy}
\end{figure}

We evaluate the effectiveness of our model by examining its accuracy in expressing real-world surface appearance both in polarimetry and radiometry. We also show that our FMBRDF can be estimated from a single polarimetric image under a known directional light.
For all experiments, input images are captured with a commercial monochrome polarization camera (Lucid TRI050S-PC) that uses quad-Bayer polarization filter chips (Sony IMX250MZR).

\subsection{Polarimetric Model Accuracy}

We first evaluate the polarimetric accuracy of FMBRDF using polarimetric images of real surfaces with known surface normals. In particular, we use spheres captured under various lighting conditions and evaluate the accuracy of FMBRDF and other existing pBRDF models in fitting the DoLP values. \Cref{fig:BRDFestimation} shows the DoLP values on the surface as a function of the angle between the global surface normal and the viewing direction for the input polarization observation, our model, and representative physically-based pBRDF models. It clearly shows that our FMBRDF model accurately explains the DoLP for all surface normals for various materials of different surface roughnesses. Other pBRDF models result in large errors for rougher surfaces as they do not model the microfacet body reflection.

\subsection{Radiometric Model Accuracy}

\begin{figure}[t]
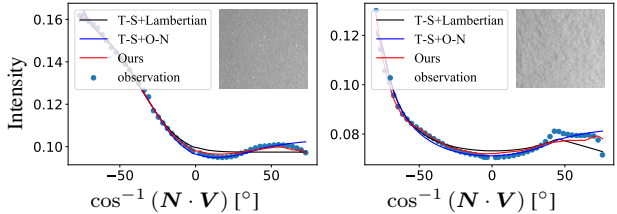

    \centering

    \includetikzgraphics[ONfitting]{fig/tikz}
    
    \caption{Radiometric fitting results on objects with non-Lambertian diffuse reflectance. The microfacet correlation function enables our model to represent Oren-Nayar diffuse reflection caused by rough mesogeometry.}
    \label{fig:radiometric accuracy O-N}
\end{figure}

We evaluate the radiometric accuracy of our FMBRDF model by fitting our model and others to images of real surfaces. \Cref{fig:radiometric accuracy} shows the results of our FMBRDF and other radiometric BRDF models on several images taken under different light source directions. These results clearly show that a Gaussian microfacet distribution is hardly appropriate for real-world surfaces. Furthermore, our FMBRDF model can represent observed body reflection more accurately than the Lambertian and Oren-Nayar models.

Our FMBRDF model subsumes the Oren-Nayar diffuse reflection model as a special case when the concentration parameter $\kappa$ of the microfacet correlation function is large. 
We capture a planar sample with rough mesogeometry by rotating the camera around it. The global surface normal $\vN$, the viewing direction $\vV$, and the light source direction $\vL$ are on the same plane. We compute the average intensity over all pixels within the surface region shared across all images. \Cref{fig:radiometric accuracy O-N} shows the fitting results of our FMBRDF and other radiometric BRDFs. The results show that our model can represent non-Lambertian diffuse reflection including Oren-Nayar diffuse reflection, in particular with its microfacet correlation function.

\subsection{Polari-Radiometric Reflectometry Accuracy}

\begin{figure}[t]
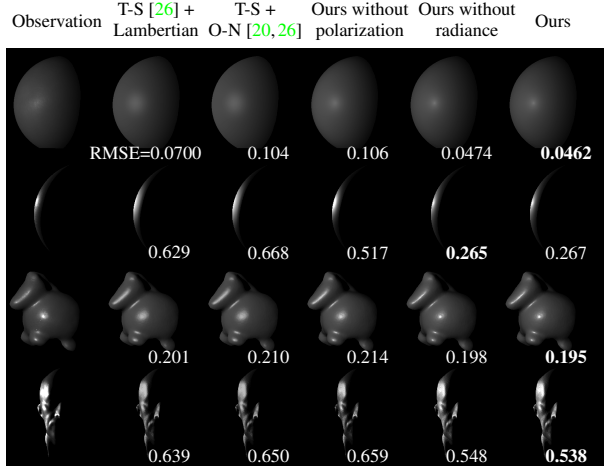

\centering
    \includetikzgraphics[reflectometry]{fig/tikz}
    \caption{Radiometric renderings using BRDF parameters estimated from single polarimetric images (ours), DoLP images (ours without radiance), and radiometric images (others). The number under the rendered image is the RMSE from the observation divided by the average of observed radiance. In contrast to BRDFs estimated with only radiance, FMBRDF estimated with radiance and polarization results in accurate radiometric appearance both in surface and body reflection for different surfaces.}
    \label{fig:renderedimage}

\end{figure}

Now, we estimate our FMBRDF from a single polarimetric image of an object with known geometry by solving \cref{eq:optimization}. We quantitatively evaluate the accuracy of the estimated FMBRDF by rendering a radiometric image under a novel lighting condition. \Cref{fig:renderedimage} shows rendered images under novel lighting conditions and prediction errors of FMBRDF and other physically-based BRDF models for objects of different color, surface roughness, and shape. The parameters of the other models and ours without polarization are estimated from the observed radiance values. Ours without radiance is estimated from only DoLPs. Radiometric BRDF models, including ours when estimated from observed radiance, fail to accurately predict the surface reflection. In contrast, images rendered with FMBRDF using parameter estimates from polarization are accurate, especially near the highlights regardless of the albedo ratio, roughness, and shape of the object. These results clearly demonstrate the accuracy of polari-radiometric reflectometry and the advantage of FMBRDF which enables it by unifying radiometric and polarimetric reflection.

\section{Conclusion}

In this paper, we derived Fresnel Microfacet BRDF, a novel analytical reflectance model that is physically accurate and generalizes to various real-world surfaces. FMBRDF models body and surface reflection with the same microgeometry represented with an oriented distribution of mirror microfacets and also unifies radiometric and polarimetric behaviors of the reflections in a single model. This resolves the physical implausibility of traditional diffuse plus specular reflection models widely adopted in computer vision and enables accurate and robust reflectance estimation for a wide range of real-world surfaces. We believe FMBRDF will serve as a sound, versatile reflectance model for radiometric and polarimetric image understanding.   

\paragraph*{Acknowledgement}
This work was in part supported by
JSPS %KAKENHI
20H05951, % 学変
21H04893, % 基板A
and JST JPMJCR20G7. % 日独仏

%%%%%%%%% REFERENCES
{\small
\bibliographystyle{ieee_fullname}
\bibliography{arxiv}

\begin{thebibliography}{10}\itemsep=-1pt

\bibitem{Atkinson2007ShapeEU}
Gary~A. Atkinson and Edwin~R. Hancock.
\newblock {S}hape {E}stimation {U}sing {P}olarization and {S}hading from {T}wo
  {V}iews.
\newblock {\em IEEE TPAMI}, 29, 2007.

\bibitem{Baek2018SimultaneousAO}
Seung-Hwan Baek, Daniel~S. Jeon, Xin Tong, and Min~H. Kim.
\newblock {S}imultaneous {A}cquisition of {P}olarimetric {SVBRDF} and
  {N}ormals.
\newblock {\em ACM Transactions on Graphics (TOG)}, 37:1 -- 15, 2018.

\bibitem{baek2020image}
Seung-Hwan Baek, Tizian Zeltner, Hyunjin Ku, Inseung Hwang, Xin Tong, Wenzel
  Jakob, and Min~H Kim.
\newblock Image-based acquisition and modeling of polarimetric reflectance.
\newblock {\em ACM Trans. Graph.}, 39(4):139, 2020.

\bibitem{ChenECCV2020}
Zhe Chen, Shohei Nobuhara, and Ko Nishino.
\newblock {I}nvertible {N}eural {BRDF} for {O}bject {I}nverse {R}endering.
\newblock In {\em Proceedings of European Conference on Computer Vision}, 2020.

\bibitem{cook1982reflectance}
Robert~L Cook and Kenneth~E. Torrance.
\newblock {A} {R}eflectance {M}odel for {C}omputer {G}raphics.
\newblock {\em ACM Transactions on Graphics (ToG)}, 1(1):7--24, 1982.

\bibitem{DanaBTF}
Kristin Dana, Bram van Ginneken, Shree~K. Nayar, and Jan~J. Koenderink.
\newblock {R}eflectance and texture of real-world surfaces.
\newblock {\em ACM Transactions on Graphics}, 18(1):1--34, Jan. 1999.

\bibitem{edwards2006halfway}
Dave Edwards, Solomon Boulos, Jared Johnson, Peter Shirley, Michael Ashikhmin,
  Michael Stark, and Chris Wyman.
\newblock {T}he {H}alfway {V}ector {D}isk for {BRDF} {M}odeling.
\newblock {\em ACM Transactions on Graphics (TOG)}, 25(1):1--18, 2006.

\bibitem{Heitz2014microfacet}
Eric Heitz.
\newblock {U}nderstanding the {M}asking-{S}hadowing {F}unction in
  {M}icrofacet-{B}ased {BRDF}s.
\newblock {\em Journal of Computer Graphics Techniques (JCGT)}, 3(2):48--107,
  June 2014.

\bibitem{hwang2022sparse}
Inseung Hwang, Daniel~S Jeon, Adolfo Mu{\~n}oz, Diego Gutierrez, Xin Tong, and
  Min~H Kim.
\newblock Sparse ellipsometry: portable acquisition of polarimetric svbrdf and
  shape with unstructured flash photography.
\newblock {\em ACM Transactions on Graphics (TOG)}, 41(4):1--14, 2022.

\bibitem{hyde2009geometrical}
Milo~W Hyde~IV, Jason~D Schmidt, and Michael~J Havrilla.
\newblock A geometrical optics polarimetric bidirectional reflectance
  distribution function for dielectric and metallic surfaces.
\newblock {\em Optics express}, 17(24):22138--22153, 2009.

\bibitem{DBLP:journals/corr/KingmaB14}
Diederik~P. Kingma and Jimmy Ba.
\newblock Adam: {A} method for stochastic optimization.
\newblock In Yoshua Bengio and Yann LeCun, editors, {\em 3rd International
  Conference on Learning Representations, {ICLR} 2015, San Diego, CA, USA, May
  7-9, 2015, Conference Track Proceedings}, 2015.

\bibitem{Koenderink98phenomenologicaldescription}
Jan~J. Koenderink and Andrea J.~Van Doorn.
\newblock Phenomenological description of bidirectional surface reflection.
\newblock {\em JOSA A}, 15:2903--2912, 1998.

\bibitem{kondo2020accurate}
Yuhi Kondo, Taishi Ono, Legong Sun, Yasutaka Hirasawa, and Jun Murayama.
\newblock Accurate polarimetric brdf for real polarization scene rendering.
\newblock In {\em European Conference on Computer Vision}, pages 220--236.
  Springer, 2020.

\bibitem{Lafortune1997}
Eric~PF Lafortune, Sing-Choong Foo, Kenneth~E Torrance, and Donald~P Greenberg.
\newblock {N}on-{L}inear {A}pproximation of {R}eflectance {F}unctions.
\newblock In {\em Proceedings of the 24th annual conference on Computer
  graphics and interactive techniques}, pages 117--126, 1997.

\bibitem{Lambert1760}
J.H. Lambert.
\newblock {\em {P}hotometria sive de mensura de gratibus luminis colorum et
  umbrae}.
\newblock Augsburg, 1760.

\bibitem{ma2007rapid}
Wan-Chun Ma, Tim Hawkins, Pieter Peers, Charles-Felix Chabert, Malte Weiss, and
  Paul~E Debevec.
\newblock {R}apid {A}cquisition of {S}pecular and {D}iffuse {N}ormal {M}aps
  from {P}olarized {S}pherical {G}radient {I}llumination.
\newblock {\em Rendering Techniques}, 2007(9):10, 2007.

\bibitem{matusik2003data}
Wojciech Matusik.
\newblock {\em {A} {D}ata-{D}riven {R}eflectance {M}odel}.
\newblock PhD thesis, Massachusetts Institute of Technology, 2003.

\bibitem{Nic77}
F.E. Nicodemus, J.C. Richmond, J.J. Hsia, W.I. Ginsberg, and T. Limperis.
\newblock {G}eometrical {C}onsiderations and {N}omenclature for {R}eflectance.
\newblock {\em Applied Optics}, 9:1474--1475, 1977.

\bibitem{nishino2009directional}
Ko Nishino.
\newblock {D}irectional {S}tatistics {BRDF} {M}odel.
\newblock In {\em 2009 IEEE 12th International Conference on Computer Vision},
  pages 476--483. IEEE, 2009.

\bibitem{Oren1995}
Michael Oren and Shree~K. Nayar.
\newblock {G}eneralization of the {L}ambertian {M}odel and {I}mplications for
  {M}achine {V}ision.
\newblock {\em International Journal of Computer Vision}, 14:227--251, 1995.

\bibitem{Phong1975IlluminationFC}
Bui~Tuong Phong.
\newblock {I}llumination for {C}omputer {G}enerated {P}ictures.
\newblock {\em Communications of the ACM}, 18:311 -- 317, 1975.

\bibitem{ramamoorthi2001signal}
Ravi Ramamoorthi and Pat Hanrahan.
\newblock A {S}ignal-{P}rocessing {F}ramework for {I}nverse {R}endering.
\newblock In {\em Proceedings of the 28th annual conference on Computer
  graphics and interactive techniques}, pages 117--128, 2001.

\bibitem{romeiro2008passive}
Fabiano Romeiro, Yuriy Vasilyev, and Todd Zickler.
\newblock {P}assive {R}eflectometry.
\newblock In {\em European Conference on Computer Vision}, pages 859--872.
  Springer, 2008.

\bibitem{smith1967geometrical}
Bruce Smith.
\newblock Geometrical shadowing of a random rough surface.
\newblock {\em IEEE transactions on antennas and propagation}, 15(5):668--671,
  1967.

\bibitem{Smith2018HfP}
William A.~P. Smith, Ravi Ramamoorthi, and Silvia Tozza.
\newblock {H}eight-from-{P}olarisation with {U}nknown {L}ighting or {A}lbedo.
\newblock {\em IEEE TPAMI}, 41(12):2875--2888, 2019.

\bibitem{torrance1967theory}
Kenneth~E Torrance and Ephraim~M Sparrow.
\newblock {T}heory for {O}ff-{S}pecular {R}eflection {F}rom {R}oughened
  {S}urfaces.
\newblock {\em Josa}, 57(9):1105--1114, 1967.

\bibitem{Walter2007microfacet}
Bruce Walter, Stephen~R. Marschner, Hongsong Li, and Kenneth~E. Torrance.
\newblock {M}icrofacet {M}odels for {R}efraction through {R}ough {S}urfaces.
\newblock In {\em Proceedings of the 18th Eurographics Conference on Rendering
  Techniques}, EGSR'07, pages 195--206. Eurographics Association, 2007.

\bibitem{Zhu2019polrgbstereo}
Dizhong Zhu and William A.~P. Smith.
\newblock {D}epth from a polarisation + rgb stereo pair.
\newblock In {\em CVPR}, 2019.

\end{thebibliography}
}

\end{document}